\documentclass[a4paper]{article}
\usepackage{graphicx}
\usepackage{onecolpceurws}
\usepackage{tikz}
\usepackage{booktabs}
\usepackage{amsmath}
\usepackage{amsfonts}
\title{Deep Learning as a Tool to Predict Flow Patterns in Two-Phase Flow}

\author{
Mohammadmehdi Ezzatabadipour  \\ University of Houston\\
                Houston, Texas 77004 \\ mezzatab@central.uh.edu
\and
Parth Singh\\ University of Houston\\
                Houston, Texas 77004 \\ psingh@central.uh.edu
\and
Melvin D. Robinson \\University of Texas at Tyler\\
                Tyler, Texas 75579 \\ mrobinson@uttyler.edu
 \and
 Pablo Guill\'en-Rondon\\ University of Houston\\
                Houston, Texas 77004 \\ pgrondon@uh.edu
 \and
 Carlos Torres\\ University of Los Andes\\
                Merida 5101,Venezuela \\ ctorres@ula.ve
}

\institution{}

\begin{document}
\maketitle
\def\layersep{2.5cm}

\begin{abstract}
In order to better model complex real-world data such as multiphase flow, one approach is to develop pattern recognition techniques and robust features that capture the relevant information. In this paper, we use deep learning methods, and in particular employ the multilayer perceptron, to build an algorithm that can predict flow pattern in two-phase flow from fluid properties and pipe conditions. The preliminary results show excellent performance when compared with classical methods of  flow pattern prediction. 
\end{abstract}
\vskip 32pt

\section{Introduction}
The term flow pattern refers to the spatial distribution of the phases, which occur during gas-liquid two-phase flow in pipes. When gases and liquids flow simultaneously in a pipe, the two phases can distribute themselves in a variety of flow configurations. The flow configurations differ from each other in the interface distribution, resulting in different flow characteristics.

Determination of flow patterns is a fundamental problem in two-phase flow analysis. Indeed all the design variables, namely, phase velocity, pressure drop, liquid holdup, heat and mass transfer coefficients, residence time distribution, and rate of chemical reaction, are all strongly dependent on the existing flow pattern. Thus, knowledge of the existing flow pattern can help the industry carry out a better design of two-phase flow systems \cite{pereyra2012methodology}. 

There is not agreement in the number of flow patterns in two-phase flow due to overlapping and characterization subjectivity, especially at the transition zones. Shoham \cite{shoham2006mechanistic} attempted to summarize the main flow patterns for all inclination angles as Dispersed bubble, Bubble, Slug, Churn, Annular and Stratified (smooth and wavy). The flow patterns depend on parameters such as pipe inclination and diameter, physical properties of the phases, and their superficial velocities. There are many models and approaches used to predict the two-phase flow patterns in pipes based on mechanistic modeling or dimensionless analysis \cite{shippen2012steady}.

Artificial neural networks have been used to identify and predict flow patterns \cite{al2016artificial}. Because of the increase in computing power researchers have developed deep learning techniques which originated from artificial neural networks. Multilayer perceptron with many hidden layers is a good example of the models with deep architectures \cite{lecun2015deep}. Deep learning techniques have been applied to a wide variety of problems in recent years \cite{langkvist2014review,yu2011deep,yuste2015neuron}. In many of these applications, algorithms based on deep learning have surpassed the previous state-of-art performance. At the heart of all deep learning algorithms is the domain independent idea of using hierarchical layers of learned abstraction to efficiently accomplish a high-level task. Deep learning allows computational models that are composed of multiple processing layers to learn representations of data with multiple levels of abstraction.

The rest of this paper is organized as follows: Section 2 explains the concept of deep learning, followed by a description of the two-phase flow patterns data base under study, and the strategy of supervised classification used in machine learning.  Section 3 present and discusses the results of the deep learning model developed in this work and finally Section 4 presents the conclusions.

\section{Materials and Methods}
The concept of deep learning originated from artificial neural network research. Unlike the neural networks of the past, modern deep learning methods have cracked the code for training stability, generalization and scale on big data.  They are often the algorithm of choice for highest predictive accuracy, as they perform well in a number of diverse problems. There are several types of learning machines for deep learning.  In our research we use a feedforward neural network known as a multilayer perceptron (MLP), a feed-forward neural network consisting of an input layer, one or more hidden layers and an output layer. Each layer is comprised of small units known as neurons. Neurons in the input layer receive the input signals $X$ and distribute them forward to the rest of the network. In subsequent layers, each neuron receives a signal, which is a weighted sum of the outputs of the nodes in the previous layer and a constant term called a bias. Inside each neuron, a nonlinear activation function transforms this input and passes it to the next layer. The advantage of such a network is that we can more accurately represent a richer set of data due to the non-linear mapping from an input vector to the output vector.  

The first step in using a multilayer perceptron is determining an appropriate architecture.  This means that we must determine the number of hidden units, the number of hidden layers, the type of activation function, as well as the number of input and output variables.  Some of this information can be determined by the structure of the training set, and some must be determined experimentally.   

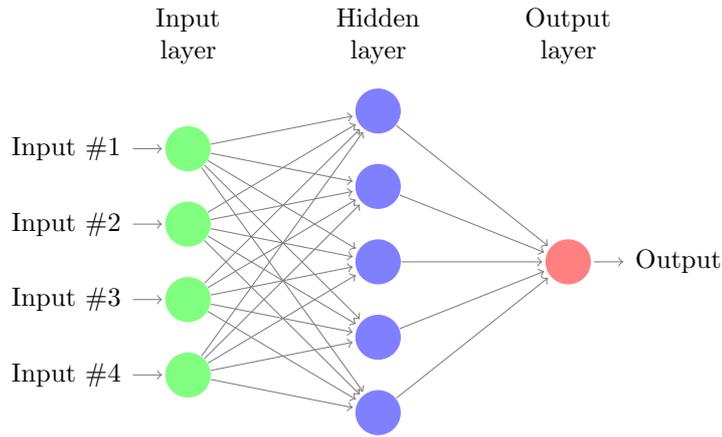
\begin{figure}
\centering
\begin{tikzpicture}[shorten >=1pt,->,draw=black!50, node distance=\layersep]
    \tikzstyle{every pin edge}=[<-,shorten <=1pt]
    \tikzstyle{neuron}=[circle,fill=black!25,minimum size=17pt,inner sep=0pt]
    \tikzstyle{input neuron}=[neuron, fill=green!50];
    \tikzstyle{output neuron}=[neuron, fill=red!50];
    \tikzstyle{hidden neuron}=[neuron, fill=blue!50];
    \tikzstyle{annot} = [text width=4em, text centered]

    \foreach \name / \y in {1,...,4}
        \node[input neuron, pin=left:Input \#\y] (I-\name) at (0,-\y) {};

    \foreach \name / \y in {1,...,5}
        \path[yshift=0.5cm]
            node[hidden neuron] (H-\name) at (\layersep,-\y cm) {};

    \node[output neuron,pin={[pin edge={->}]right:Output}, right of=H-3] (O) {};

    \foreach \source in {1,...,4}
        \foreach \dest in {1,...,5}
            \path (I-\source) edge (H-\dest);

    \foreach \source in {1,...,5}
        \path (H-\source) edge (O);

    \node[annot,above of=H-1, node distance=1cm] (hl) {Hidden layer};
    \node[annot,left of=hl] {Input layer};
    \node[annot,right of=hl] {Output layer};
\end{tikzpicture}
\caption{Illustration of a single layer multilayer perceptron}
\label{fig:f1}
\end{figure}

The second step involves determining a training algorithm to estimate the weights and biases of the MLP. This involves iteratively solving an optimization problem to estimate the network's weights and biases so that the network's output is as close to a desired output as possible. The structure of a MLP network is shown in Figure \ref{fig:f1}. 

\subsection{Dataset}
A flow pattern experimental data base was collected \cite{pereyra2012methodology}, which consists of the most relevant studies developed in the area. Specifically for this study, the data set from Shoham \cite{shohamflow} was selected among the available sets due to its large number of data points (5676), range in inclination angle ($-90^\circ$ to $90^\circ$), two pipe diameters (ID=1in and 2in), and the wide range of flow patterns observed for all pipe inclination angles. The flow patterns considered in this study are: Annular (A), Bubble (B), Dispersed bubble (DB), Intermittent (I), Stratified smooth (SS) and Stratified wavy (SW). The Intermittent flow pattern considers Slug (SL) and Churn (CH) flow pattern combined \cite{pereyra2012methodology}. In order to analyze the performance of the algorithm, three tests are proposed: Test 1 considers all the flow patterns proposed; Test 2 combines the SS and SW data points into stratified flow ST (ST = SS + SW); finally Test 3 combines the segregated flow patterns (ST + A) and the dispersed flow patterns (DB + B).
\subsection{Supervised Classification of Flow Pattern Using Machine Learning}
In supervised learning we assume each element of study is represented as an n-component vector-valued random variable, $(X_1,X_2,\dots,X_n)$ where each $X_i$  represents an attribute or feature; the space of all possible feature vectors is called the input space $X$. We also consider a set $\{y_1,y_2,\dots,y_k\}$ corresponding to the $k$ possible classes; this forms the output space $Y$.  A classifier or learning algorithm typically receives a set of training examples from a source domain $T=(\mathbf x,\mathbf y)$, where $\mathbf x=(x_1,x_2,\dots,x_n)$ is a vector in the input space, and $\mathbf y\in \mathbb{R}^k$ is a value in the (discrete) output space. We assume the training or source sample $T$ consists of independently and identically distributed (i.i.d.) examples obtained according to a fixed but unknown joint probability distribution, $P(\mathbf x,\mathbf y)$, in the input-output space.  The output of the classifier is a hypothesis or function $f(\mathbf x)$ mapping the input space to the output space, $f:X\rightarrow Y$.  In supervised learning, this mapping is the result of optimizing a loss function where our ultimate goal is to minimize number of misclassifications.  For this research we use the Python bindings of the H2O library \cite{arora2015deep} as our deep learning platform.
\section{Results}
In our approach, we trained a MLP on a set of randomly selected samples, approximately $60\%$ of the entire dataset was used for training, approximately $20\%$ was used for validation, and approximately $20\%$ was used as the testing set for the $3$ different tests. Using the multilayer perceptron we can train using simple stochastic gradient descent \cite{bottou2010large}. Table \ref{tb:arch} shows our chosen multilayer perceptron architecture and parameters used in our experiments.  Other design considerations are the choice of ReLU as our non-linear activation function, the $\ell_1$ and $\ell_2$ regularization penalties to avoid overfitting, the number of epochs, and nfolds, the number of cross-validation folds.

\begin{table}[!h]
\centering
\caption{Multilayer Perceptron architecture and parameters}
\bigskip

\label{tb:arch}
\begin{tabular}{lc}
\toprule
Variables& Parameters\\\midrule
Number of input neurons&$11$\\
Number of hidden layers&$3$\\
Hidden layer topology&$(25,25,25)$\\
Number of output neurons (classes)&$6$\\
Activation function&ReLU\\
Loss function &Mean Squared Error\\
Number of training epochs& 1000000\\
$\ell_1$ penalty weighting &$0.00001$\\
$\ell_2$ penalty weighting &$0.00001$\\
n-fold cross-validation & $10$\\
\bottomrule
\end{tabular}
\end{table}



In evaluating the effectiveness of our deep learning methodology, the confusion matrix is an important measure. Table \ref{tb:conf1} shows the confusion matrix for the training data set for Test 1, predicting classes A, B, DB, I, SS, and SW.  We can readily see the strong diagonal components.  This means that our classifier is achieving little classification error.  The testing set is used to predict the variable Flow Pattern, which contains labels for each class (A, B, DB, I, SS, and SW), and a predictive accuracy of $83.87\%$ for the different classes is obtained, the details of which are shown in Table \ref{tb:conf_val_t1}.  Off diagonal elements of a confusion matrix show misclassification with other flow patterns.  The confusion matrix's columns represent the output patterns predicted by Deep Learning while the rows represent the true class which is denoted here by each flow pattern.


\begin{table}[h]
\centering
\caption{Training Data Confusion Matrix: Test 1}

\bigskip

\label{tb:conf1}
\begin{tabular}{c c c c c c c c c}
\toprule
&A & B&DB & I & SS&SW&Error&Rate\\\midrule
A&617&0&0&4&0&0&0.0064412&4/621\\
B&0&76&0&0&0&0&0.0&0/76\\
DB&0&0&331&34&0&0&0.0931507&34/365\\
I&42&46&60&1629&0&6&0.0863713&154/1783\\
SS&0&0&0&56&14&10&0.825&66/80\\
SW&114&0&1&43&5&330&0.3306288&163/493\\\midrule
&773&122&392&1766&19&346&0.1231714&421/3418\\\bottomrule
\end{tabular}
\end{table}

\begin{table}
\centering
\caption{Confusion matrix for the cross-validation data set Test 1}

\bigskip

\label{tb:conf_val_t1}
\begin{tabular}{ccccccccc}\toprule
&A&B&DB&I&SS&SW&Error&Rate\\\midrule
A&581&0&3&20&0&17&0.0644122&40/621\\
B&0&51&0&25&0&0&0.3289474&25/76\\
DB&2&0&310&50&0&3&0.1506849&55/365\\
I&78&29&65&1474&67&70&0.1733034&309/1783\\
SS&0&0&0&5&67&8&0.1625&13/80\\
SW&94&0&2&23&26&348&0.2941176&145/493\\\midrule
&755&80&380&1597&160&446&0.1717379&587/3418\\\bottomrule
\end{tabular}
\end{table}
Table \ref{tb:conf2} shows the confusion matrix on train data for Test 2, predicting classes A, B, DB, I and ST.  Similar to Test 1 we can see the strong diagonal components, and the classifier has small classification error.  The testing set is used to predict the variable Flow Pattern, which contains labels for each class (A, B, DB, I, and ST), and we achieve a predictive accuracy of $83.34\%$.  The details are shown in Table \ref{tb:conf_val_2}.


\begin{table}[ht]
\centering
\caption{Training Data Confusion Matrix: Test 2}

\bigskip

\label{tb:conf2}
\begin{tabular}{cc c c c c c c}
\toprule
&A & B&DB & I & ST&Error&Rate\\\midrule
A&605&0&0&6&10&0.0257649&16/621\\
B&0&76&0&0&0&0&0/76\\
DB&0&0&350&14&1&0.2&15/365\\
I&69&40&109&1433&132&0.1962984&350/1783\\
ST&93&0&1&1&478&0.1657941&95/573\\\midrule
&767&116&460&1454&621&0.1392627&476/3418\\\bottomrule
\end{tabular}
\end{table}


\begin{table}[ht]
\centering
\caption{Confusion matrix for the cross-validation data set Test 2}

\bigskip

\label{tb:conf_val_2}
\begin{tabular}{cccccccc}\toprule
&A&B&DB&I&ST&Error&Rate\\\midrule
A&574&0&2&21&24&0.0756844 &47/621\\
B&0& 35& 1&40&0&0.5394737 &41/76\\
DB&0&0&309&53&3&0.1534247 &56/365\\
I&82&15.0&94.0&1436.0&156.0&0.1946158&347/1783\\
ST&98&0.0&4.0&28.0&443.0&0.2268761&130/573\\\midrule
&754&50&410&1578&626&0.1816852&621/3418\\\bottomrule
\end{tabular}
\end{table}



%
%


\begin{table}[ht]
\centering
\caption{Confusion matrix for the cross-validation data set Test 3}

\bigskip

\label{tb:conf_train_3}
\begin{tabular}{lccccc}\toprule
&Intermittent&Dispersed&Segregate&Error&Rate\\\midrule
Intermittent&1419&63&301&0.2041503&364/1783\\
Dispersed&64&295&6&0.1917808&70/365\\
Segregate&85&2&1183&0.0685039&87/1270\\\midrule
&1490&420&1508&0.1524283&521/3418\\\bottomrule
\end{tabular}
\end{table}

\begin{table}[ht]
\centering
\caption{Confusion matrix for the cross-validation data set Test 3}

\bigskip

\label{tb:conf_val3}
\begin{tabular}{cccccc}\toprule
&Intermittent&Dispersed&Segregate&Error&Rate\\\midrule
Intermittent&1469&70&244&0.1761077&314/1783\\
Dispersed&14&350&1&0.0410959&15/365\\
Segregate&7&0&1263&0.00055118&7/1270\\\midrule
&1490&420&1508&0.0983031&336/3418\\\bottomrule
\end{tabular}
\end{table}
%
%

Table \ref{tb:conf_train_3} shows the confusion matrix for the training data set for Test 3, predicting the classes Intermittent, Dispersed, and Segregate.  We can readily see the strong diagonal components with the classifier achieving little classification error.  The testing set is used to predict the variable flow pattern, which contains labels for each class (Intermittent, Dispersed, and Segregate) with a predictive accuracy of $85.97\%$, the details of which are shown in Table \ref{tb:conf_val3}

A comparison between the predicted flow pattern and the experimental database considering the three data sets under study show low error and high classification accuracy.  Results for Test 1 and Test 2 are very similar.  Most of the failed predictions between the flow patterns can be attributed to the different criteria used by the different experimentalists to classify the flow patterns and their relationships \cite{shohamflow}.  Finally an improvement is obtained for Test 3 by combining the segregated flow patterns (ST + A) and the dispersed flow patterns (DB + B).  The prediction accuracy for this case increases to $85.97\%$.  This is improvement is due to the clear and straightforward distinction between the two combined flow patterns \cite{pereyra2012methodology,shohamflow}.  The results for the deep learning approach for classification of two-phase flow pattern are encouraging. 

\section{Conclusions}
In this paper we proposed three types of data sets as input features and investigated the use of deep learning for the classification and prediction of two-phase flow, based on experimental data, obtained from \cite{pereyra2012methodology,shohamflow}.

We proposed six types of input features, and a corresponding architecture to precisely predict flow patterns. First, we showed that the network can learn surprisingly well as using our chosen architecture and parameters allowed us to achieve high classification accuracy. Second, we showed that the network can classify the different flow patterns with high efficiency. Finally, we achieved high precision predicting different combinations of classes. 

Our experiments indicate that a deep learning approach, has the potential to capture flow patterns, which may boost the classification performance. These investigations could be further improved in future studies by carrying out more exhaustive searches for the parameters in the architectures. The result would be improved overall performance of these systems.

Finally, deep learning can be used to predict flow patterns using pipe characteristics, fluid properties and superficial velocities of the two-phase flows. It outperforms results from previous studies.\cite{pereyra2012methodology}

\newpage

\bibliographystyle{alpha} 
\bibliography{dm4og}

\newcommand{\etalchar}[1]{$^{#1}$}
\begin{thebibliography}{ANEAS16}

\bibitem[ACL{\etalchar{+}}15]{arora2015deep}
Anisha Arora, Arno Candel, Jessica Lanford, Erin LeDell, and Viraj Parmar.
\newblock {Deep Learning with H2O}, 2015.

\bibitem[ANEAS16]{al2016artificial}
Mustafa Al-Naser, Moustafa Elshafei, and Abdelsalam Al-Sarkhi.
\newblock Artificial neural network application for multiphase flow patterns
  detection: A new approach.
\newblock {\em Journal of Petroleum Science and Engineering}, 145:548--564,
  2016.

\bibitem[Bot10]{bottou2010large}
L{\'e}on Bottou.
\newblock Large-scale machine learning with stochastic gradient descent.
\newblock In {\em Proceedings of COMPSTAT'2010}, pages 177--186. Springer,
  2010.

\bibitem[LBH15]{lecun2015deep}
Yann LeCun, Yoshua Bengio, and Geoffrey Hinton.
\newblock Deep learning.
\newblock {\em Nature}, 521(7553):436--444, 2015.

\bibitem[LKL14]{langkvist2014review}
Martin L{\"a}ngkvist, Lars Karlsson, and Amy Loutfi.
\newblock A review of unsupervised feature learning and deep learning for
  time-series modeling.
\newblock {\em Pattern Recognition Letters}, 42:11--24, 2014.

\bibitem[PTM{\etalchar{+}}12]{pereyra2012methodology}
E~Pereyra, C~Torres, R~Mohan, L~Gomez, G~Kouba, and O~Shoham.
\newblock A methodology and database to quantify the confidence level of
  methods for gas--liquid two-phase flow pattern prediction.
\newblock {\em Chemical Engineering Research and Design}, 90(4):507--513, 2012.

\bibitem[SB12]{shippen2012steady}
Mack Shippen and William~J Bailey.
\newblock {Steady-State Multiphase Flow--Past, Present, and Future, with a
  Perspective on Flow Assurance}.
\newblock {\em Energy \& Fuels}, 26(7):4145--4157, 2012.

\bibitem[Sho]{shohamflow}
O~Shoham.
\newblock {Flow Pattern Transition and Characterization in Gas-Liquid Two-Phase
  Flow in Inclined Pipes. 1982}.
\newblock {\em PhD Dissertation}.

\bibitem[Sho06]{shoham2006mechanistic}
Ovadia Shoham.
\newblock {\em Mechanistic modeling of gas-liquid two-phase flow in pipes}.
\newblock Richardson, TX: Society of Petroleum Engineers, 2006.

\bibitem[YD11]{yu2011deep}
Dong Yu and Li~Deng.
\newblock Deep learning and its applications to signal and information
  processing [exploratory dsp].
\newblock {\em IEEE Signal Processing Magazine}, 28(1):145--154, 2011.

\bibitem[Yus15]{yuste2015neuron}
Rafael Yuste.
\newblock From the neuron doctrine to neural networks.
\newblock {\em Nature Reviews Neuroscience}, 16(8):487--497, 2015.

\end{thebibliography}

%
%
%

\end{document}